\pdfoutput=1

\documentclass[11pt]{article}

\usepackage[final]{acl}

\usepackage{times}
\usepackage{latexsym}

\usepackage[T1]{fontenc}

\usepackage[utf8]{inputenc}

\usepackage{microtype}

\usepackage{inconsolata}

\usepackage{graphicx}

\usepackage{amsmath}
\usepackage{amsthm}
\usepackage{booktabs}
\usepackage[switch]{lineno}
\usepackage[ruled,vlined,linesnumbered]{algorithm2e}
\SetKw{KwTo}{to}
\SetKw{KwBy}{by}
\usepackage{url}
\usepackage{subfigure}

\definecolor{aureolin}{rgb}{0.99, 0.93, 0.0}
\usepackage{multirow}
\definecolor{lightgreen}{RGB}{217,255,179}

\newcommand{\medvoc}{\textsc{MEDVOC}}
\newcommand{\drift}{\textsc{MEDVOC}}
\newcommand{\base}{\textsc{PLM}}
\newcommand{\adapt}{\textsc{AdaptBPE}}
\newcommand{\bpe}{\textsc{BPE}}
\newcommand{\avocado}{\textsc{AVocaDo}}
\newcommand{\vdomain}{V\textsubscript{DOMAIN}}

\usepackage{amssymb}%
\usepackage{pifont}%

\title{Adaptive BPE Tokenization for Enhanced Vocabulary Adaptation in Finetuning Pretrained Language Models}

\author{
  \textbf{Gunjan Balde}\textsuperscript{\textsection},
  \textbf{Soumyadeep Roy}\textsuperscript{\textsection},
  \textbf{Mainack Mondal}
  and \textbf{Niloy Ganguly}
\\
Indian Institute of Technology Kharagpur 
\\
\texttt{balde.gunjan0812@kgpian.iitkgp.ac.in}\\
\texttt{soumyadeep.roy9@iitkgp.ac.in} \\
\texttt{\{mainack,niloy\}@cse.iitkgp.ac.in}\\
}

\begin{document}
\maketitle
\begingroup\renewcommand\thefootnote{\textsection}
\footnotetext{Equal Contribution. Corresponding author: \texttt{balde.gunjan0812@kgpian.iitkgp.ac.in}. Accepted for publication at The 2024 Conference on Empirical Methods in Natural Language Processing (EMNLP Findings). This is the author’s version of the work. It is posted here for your personal use. Not for redistribution.}

\begin{abstract}
In this work, we show a fundamental limitation in vocabulary adaptation approaches that use Byte-Pair Encoding (\bpe) tokenization scheme for fine-tuning pretrained language models (PLMs) to expert domains. Current approaches trivially append the target domain-specific vocabulary (\vdomain{}) at the end of the PLM vocabulary. This approach leads to a lower priority score and causes sub-optimal tokenization in \bpe{} that iteratively uses merge rules to tokenize a given text. To mitigate this issue, we propose \adapt{} where the \bpe{} tokenization initialization phase is modified to first perform the longest string matching on the added (target) vocabulary before tokenizing at the character level. We perform an extensive evaluation of \adapt{} versus the standard BPE over various classification and summarization tasks; \adapt{} improves by $3.57\%$ (in terms of accuracy) and $1.87\%$ (in terms of Rouge-L), respectively. \adapt{} for \medvoc{} works particularly well when reference summaries have high OOV concentration or are longer in length. We also conduct a human evaluation, revealing that \adapt{} generates more relevant and more faithful summaries as compared to \medvoc. We make our codebase publicly available at \url{https://github.com/gb-kgp/adaptbpe}.

\end{abstract}

\section{Introduction}

\begin{figure}[!ht]
    \centering
    \includegraphics[width=\columnwidth]{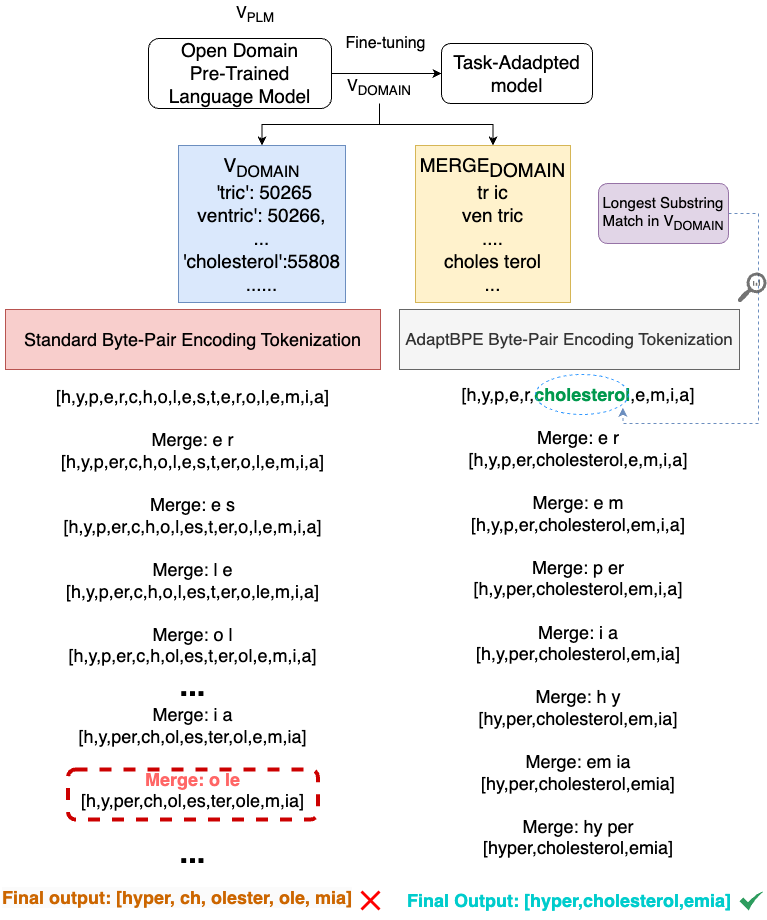}
    \caption{ \adapt{} modifies the initialization step of standard \bpe{} by merging the characters that match with the extended vocabulary (\vdomain{}). The incorrect merge step of \bpe{} for tokenizing the word \textit{hypercholesterolemia} is highlighted by a red dashed box.} 
    \label{fig:BPE-IllSplit}
\end{figure}

Vocabulary adaptation-based fine-tuning has proved successful in domain adaptation to expert domains, characterized by high vocabulary mismatch. Here, the PLM vocabulary is further extended by adding a target domain-specific vocabulary (\vdomain{}) during fine-tuning. To identify \vdomain{} works like VOLT~\cite{xu-etal-2021-vocabulary} and \avocado~\cite{hong-etal-2021-avocado} focus on optimizing the model's vocabulary by adding subwords based on utility scoring functions that are based on variants of fragment score~\cite{rust-etal-2021-good} or optimize Pointwise Mutual Information~\cite{diao-etal-2021-taming} or by measuring domain shift of token sequence distribution~\cite{sachidananda-etal-2021-efficient}. \medvoc~\cite{balde2024medvoc}, is the first work in a summarization setting that uses fragment score as the utility function. \textbf{In this work, we establish the need also to adapt the tokenization scheme}; Figure~\ref{fig:BPE-IllSplit} provides an example of ill-tokenization due to the limitations of the standard BPE tokenization scheme. 

Prior vocabulary adaptation studies~\cite{hong-etal-2021-avocado,balde2024medvoc} append added vocabulary and corresponding merge rules towards the end of existing PLM vocabulary (V\textsubscript{PLM}). \textbf{This approach does not guarantee that the Byte-Pair Encoding (\bpe) tokenizer will use the added target domain vocabulary}. We believe this is because the merge rules are trivially appended to the end, automatically implying a lower priority (of \vdomain{}) over existing PLM vocabulary (V\textsubscript{PLM}). 

Our main contribution is to propose the \adapt{} tokenization scheme that mitigates the above-mentioned ill-tokenization issue of \bpe{} when applied to vocabulary adaptation strategies. Our proposed \adapt{} algorithm is independent of the target domain-specific vocabulary construction algorithm and only modifies the underlying BPE tokenization phase. \textbf{\adapt{} modifies the initialization stage of a standard \bpe{} tokenization} as explained in detail in Algorithm~\ref{algo:bpe_tokenization}. Instead of starting tokenization by splitting the input token to character level, \adapt{} performs the longest substring match in the added vocabulary (\vdomain{}) iteratively and preserves the matched substring from splitting into characters further. This modified \bpe{} algorithm, \adapt, mitigates the ill-tokenization issues completely, as we observe a significant drop in fragment score (average number of subwords a given word across the entire corpus) of $39.16\%$ and $13.96\%$ in case of \avocado{} and \medvoc{} respectively.

\adapt{} shows improvements of $3.57\%$ and $1.87\%$ over the standard \bpe{} algorithm in the case of \avocado{} and \medvoc{} respectively for eight datasets (4 classification and 4 summarization tasks). In the case of \medvoc{} for difficult generation scenarios such as high OOV (out-of-vocabulary) concentration and longer reference summaries, \adapt{} consistently improves by $10.41\%$ and $3.30\%$ in terms of Rouge-L. We further perform a human evaluation using medical experts where we observe that \adapt{} produces more relevant and faithful summaries in the case of \medvoc. We make our codebase publicly available at \url{https://github.com/gb-kgp/adaptbpe}.

\section{Background}
\noindent \textbf{Vocabulary Adaptation Strategies for Classification --\avocado.} \avocado~\cite{hong-etal-2021-avocado} propose a vocabulary adaptation strategy for classification tasks. \avocado{} iteratively adds task-specific vocabulary (\vdomain{}) constructed from source documents of target tasks to existing PLM vocabulary (V\textsubscript{PLM}). The amount of vocabulary to be added is decided using fragment score, which is defined as the average number of subwords tokenized per word given a vocabulary. \avocado{} starts on a set of words that are split into more than two subwords ($W_{s>2}$) and constructs task-specific vocabulary on this set of words. It then keeps on adding the vocabulary from this task-specific vocabulary till the fragment score of words in set ($W_{s>2}$) stays above a fixed threshold, $\gamma$. \avocado{} initialized the embeddings of the newly added subwords with the average of embeddings of the subwords they were previously split into. \avocado{} uses contrastive loss framework~\cite{chen2020simple} as a regularization loss along with the standard cross-entropy loss for classification to tune the model with the added embeddings of the newly added subwords.

\noindent \textbf{Vocabulary Adaptation Strategies for Summarization --\medvoc.} \medvoc~\cite{balde2024medvoc} proposes a vocabulary adaptation framework for summarization tasks in the medical domain for three models -- BERT, BART, and PEGASUS. First, \medvoc{} identifies vocabulary to be added (\vdomain{}) as an optimizable parameter. It constructs vocabulary on candidate set of medical OOV (Out-Of-Vocabulary) words (words that are medical, and split into more than one word using existing PLM vocabulary) identified from combination of PAC (PubMed Abstract Collection) dataset (V\textsubscript{PAC}) and target-task specific datasets (V\textsubscript{TGT}). It then performs a hyperparameter search using fragment score as the metric, over different vocabulary sizes and identifies the optimal vocabulary to be added to existing PLM vocabulary (V\textsubscript{PLM}). The embeddings are initialized randomly and are tuned by performing an intermediate fine-tuning step on PAC dataset comprising PubMed abstract as source document and the title as reference summary.

\section{Proposed Methodology} \label{sec:methods}

\noindent \textbf{Working of the standard \bpe{} Tokenization.} \bpe{} is the most common tokenization scheme that is found to be most effective among various tokenization strategies~\cite{galle-2019-investigating,zouhar-etal-2023-formal,schmidt2024tokenization}, and is used in the majority of recent Large Language Models (LLMs) like LLaMa~\cite{touvron2023llama1,touvron2023llama2} and Mistral~\cite{jiang2023mistral}. BPE tokenization scheme takes as input two files: (i) vocabulary file, which contains the vocabulary, and (ii) merge rules file, which contains merge rules for the terms present in the vocabulary required for its construction (e.g., {\it th e} merge rule for the word \textit{`the'} in vocabulary). The standard BPE tokenizer starts by splitting the input word into the character level. Then, following a bottom-up strategy, it iteratively merges adjacent characters following the ordered merge rules from the merge rule file taken as input. For instance, consider the word \textit{happy}. BPE starts by converting this word into a list of characters: \textit{[h, a, p, p, y]}. Then, it checks for possible merges on adjacent characters and selects the one with the least rank. Here, it chooses {\it <p,p>} resulting in \textit{[h, a, pp, y]}. It then iteratively keeps checking and ends with [\textit{`happy'}] as the final output for BPE tokenization.

\vspace{1mm}
\noindent \textbf{\adapt{} Tokenization Scheme (Algorithm~\ref{algo:bpe_tokenization}).} We observe that the main reason for ill-tokenization (See Figure~\ref{fig:BPE-IllSplit}) is certain merge rules that hinder the formation of added vocabulary. Therefore, instead of splitting at the character level at the initialization stage, we first check for the longest substring match~\cite{hofmann-etal-2022-embarrassingly} only in the added vocabulary (\vdomain{}) and prevent the match from splitting into the character level. This step is iterated till we cannot find any substring match. Figure~\ref{fig:BPE-IllSplit} shows an example: the word \textit{hypercholesterolemia} is initialized as \textit{[h,y,p,e,r,cholesterol,e,m,i,a]} as opposed to standard BPE tokenization which starts entirely at character level: \textit{[h,y,p,e,r,c,h,o,l,e,s,t,e,r,o,l,e,m,i,a]}. 

\newcommand\mycommfont[1]{\tiny\ttfamily\textcolor{blue}{#1}}
\SetKwInput{KwInput}{Input}
\SetKwInput{KwOutput}{Output}
\SetKwInput{KwData}{Data}
\SetCommentSty{mycommfont}
\SetKw{Continue}{continue}
\SetKw{Break}{break}

\begin{algorithm}[!ht]
\scriptsize
\caption{\adapt{} tokenization}\label{algo:bpe_tokenization}
\DontPrintSemicolon
    \KwInput{Text text, Tokenizer tokenizer, Merge rules merges, Added vocabulary \vdomain{}}
    \KwOutput{BPE token sequence $\mathcal{T}$}

    \tcp{Pre-tokenizing the text based on pre-tokenization rules of tokenizer}
    $\text{pre\_tokenized} \gets \text{tokenizer.pre\_tokenize\_str(text)}$ \; 
        
    $\text{pre\_tokenized\_text} \gets [\text{word} \text{ for } \text{word} \text{ in } \text{pre\_tokenized}]$\;
    
    $\mathcal{T} \gets []$ \;
    
    \For{$\text{word} \in \text{pre\_tokenized\_text}$}{
        
        split $\gets$ \{\} \;

        \tcp {Finding the longest substring match in \vdomain{}}
        remaining $\gets$ word \;
        
        \While{$True$}{
        
            idx\textsubscript{match} , longest\textsubscript{match} $\gets$ longest\_substr(remaining, \vdomain{}) \;

            \If{idx\textsubscript{match} $== -1$}{
                
                \Break \;
                
            }
            
            \Else{
            
                split[idx\textsubscript{match}] $\gets$ longest\textsubscript{match} \;
                
                \For{i in range(idx\textsubscript{match}, idx\textsubscript{match}+longest\textsubscript{match}.length)}{
                remaining[i] $\gets$ `-' \;
                
                }
            }
        }

        \tcp{Retrieving the longest matches and remaining parts of string}
        
        subwords $\gets$ [sw for i, sw in sorted(split)]

        pairs $\gets$ get\_bigrams(subwords) \;
        
        \tcp{Standard BPE loop}
        
        \While{$True$}{
            bigram $\gets$ \{least ranking applicable merge rule on pairs\}  \;
            
            \If{bigram is invalid}{
                \Break
            }
        
            first, second $\gets$ bigram \;
            
            new\_word $\gets$ [] \;
            
            i := 0 \;
                        
            \While{$i < $ subwords.length}{
                    $j := $ subwords.index(first, i)  \;

                    new\_word.extend(subwords[i:j]) \;
                    
                    $i := j $ \;

                \If{subwords[i] == first and i < len(subwords) - 1 and subwords[i + 1] == second}{
                    new\_word.append(first + second) \;
                    
                    $i := i + 2$ \;
                }
                
                \Else{
                    new\_word.append(subwords[i]) \;
                    
                    $i :=  i +1$
                }
            }
            new\_word := tuple(new\_word) \;
            
            subwords := new\_word \;
            
            \If{subwords.length == 1}{
                \Break
            }
            \Else{
                pairs $\gets$ get\_bigrams(subwords) \;
            }
        }   

        $\mathcal{T}$.extend(subwords) \;
    }
    \KwRet {$\mathcal{T}$}
\end{algorithm}
\section{Experimental Setup}\label{sec:expt-setup}

We use the same experimental setup as the state-of-the-art vocabulary adaptation works of \avocado{} and \medvoc{} for the classification and summarization tasks, respectively. Appendix~\ref{appendix:expt-setup} provides all the necessary implementation details. 

\paragraph{Datasets.} We use the same datasets as used in \avocado{} and \medvoc{} (see Appendix~\ref{appendix:datasets} for further details) --- (i) four classification tasks: \textsc{CHEMPROT}~\cite{kringelum2016chemprot} from the biomedical domain, \textsc{ACL-ARC}~\cite{jurgens2018measuring} from the computer science domain, \textsc{HYPERPARTISAN} (HYP)~\cite{kiesel2019semeval} from the news domain, and \textsc{AMAZON}~\cite{amazon-mcauley-2015} from the customer reviews domain, and (ii) two query-focused document summarization datasets: EBM~\cite{DBLP:conf/acl-alta/MollaS11a} and BioASQ~\cite{cite:BioASQ}, and two question summarization datasets:  MeQSum~\cite{ben-abacha-demner-fushman-2019-summarization} and CHQSum~\cite{yadav2022chq}.

\noindent \textbf{Evaluation Metrics.}  We report classification accuracy and Macro-F1 scores for the classification task. For summarization, we report Rouge-L~\cite{lin-2004-rouge} and Concept Score ~\cite{zhang2023famesumm}, which measures the overlap of UMLS medical concepts between the generated and reference summaries. See Appendix~\ref{appendix:eval-metrics} for additional details.
\section{Performance Evaluation of \adapt} \label{sec:result_discussion}
We show the performance comparison results of \adapt{} in Table~\ref{tab:avocado} and~\ref{tab:medvoc} for the vocabulary adaptation strategies for the classification and summarization setting, respectively. Please refer Appendix~\ref{appendix:added-vocab-sizes} for the sizes of added vocabularies in both the settings for all the datasets and Appendix~\ref{appendix:hyperparameters} for relevant hyperparameter details.

\noindent \textbf{Performance Evaluation in Classification Datasets.} We show the performance comparison of BPE versus  \adapt~in Table~\ref{tab:avocado}. We observe gains of $3.57\%$ in accuracy and 3.18\% in the case of the Macro-F1 score. We further observe huge drops in fragment scores of 39.16\% across four datasets from four domains. Thus, \adapt{} helps to correctly tokenize the domain words, which leads to better performance.

\begin{table}[t]
    \centering
    \scriptsize
    \resizebox{\columnwidth}{!}{
    \begin{tabular}{p{1.2cm}cccc}
    \hline
    
      \textbf{Dataset (Domain)} & \textbf{Model}  & \textbf{FragSr} & \textbf{Accuracy} & \textbf{Macro-F1} \\ \hline
      \textsc{CHEMPROT}  & BPE & 2.55 & \textbf{81.43} $\pm$ 0.55 & 54.88 $\pm$ 1.66\\
      (BioMedical)& \adapt & \textbf{1.16} & 81.40 $\pm$ 0.40 & \textbf{55.02} $\pm$ 0.47 \\ \hline

    \textsc{ACL-ARC}\textsuperscript{*}   & BPE & 2.21 &  69.03 $\pm$ 5.05 & 55.04 $\pm$ 8.24 \\
    (Computer Science)&  \adapt &\textbf{1.18} & \textbf{73.02} $\pm$ 4.21 & \textbf{62.00} $\pm$ 4.95 \\ \hline

     \textsc{HYP}   & BPE  & 3.26  & 77.84 $\pm$ 5.20  & 74.23 $\pm$ 7.01 \\
     (News) &  \adapt     &  \textbf{3.17} &  \textbf{82.16} $\pm$ 2.50 & \textbf{80.64} $\pm$ 3.03 \\ \hline 

     \textsc{AMAZON} & BPE  & 2.81  & 83.13 $\pm$ 3.64 & 68.34 $\pm$ 0.47 \\
     (Reviews)      & \adapt & \textbf{2.47} & \textbf{86.26} $\pm$ 0.53 & \textbf{69.90} $\pm$ 0.29\\ \hline
    \end{tabular}}
    \caption{Performance evaluation of \adapt{} on \avocado{} with RoBERTa-Base as base model averaged across 5 seeds (* -except for ACL-ARC which was done for 20 seeds). Improvements wherever observed are statistically significant (t-test: p-value$<0.05$). We show improvements of $3.57\%$ in accuracy and $3.18\%$ in the case of the Macro-F1 score. \adapt~results in the fragment score (FragSr) drop of $39.16\%$ across datasets.}
    \label{tab:avocado}
\end{table}

\noindent \textbf{Performance Evaluation in Medical Summarization Datasets.} We show the performance comparison of BPE versus  \adapt in Table~\ref{tab:medvoc}. We observe gains of $1.87\%$ in Rouge-L (R-L) and $0.9\%$ in the case of ConceptScore (CSr). We further observe huge drops in fragment scores of $13.20\%$ across four datasets. This indicates the efficacy of \adapt, as we are now correctly tokenizing the words and thus the downstream task improvement. We further investigate how \adapt{} performed compared to BPE when reference summaries had high OOV concentration and were long in length following evaluation as performed in \medvoc. These points represent the most difficult data points in terms of vocabulary mismatch. \adapt{} shows a big improvement of $10.40\%$ on average across four datasets in high OOV settings and $3.41\%$ on average across BioASQ and EBM datasets for a long-form generation.

\begin{table}[!ht]
\centering
\scriptsize
\resizebox{\columnwidth}{!}{
\begin{tabular}{cccccc}
\hline

\textbf{Model} & \textbf{FragSr} & \textbf{R-L\textsubscript{All}} & \textbf{CSr\textsubscript{All}} & \textbf{R-L\textsubscript{H-O}} & \textbf{R-L\textsubscript{L-RS}}   \\ \hline

\multicolumn{6}{c}{\textbf{EBM}}  \\
BPE    & 3.00  & 20.65 & 22.66 & 19.23 & 17.62	 \\
\adapt & \textbf{2.31} & \textbf{20.73} & \textbf{22.67} & \textbf{21.43} & \textbf{17.74}   \\ \hline

\multicolumn{6}{c}{\textbf{BioASQ}} \\
BPE    & 3.14 & 48.02 & 52.87 & 39.23 & 43.25\\
\adapt & \textbf{2.71} & 47.72 & \textbf{52.93} & \textbf{42.95} & \textbf{45.91}\\ \hline

\multicolumn{6}{c}{\textbf{MeQSum}}\\ 
BPE    & 3.34 & 55.88 & 60.52  & 75.56 &  -\\ 
\adapt & \textbf{3.15} &\textbf{58.00} & \textbf{62.29} & \textbf{82.64} & -\\ \hline

\multicolumn{6}{c}{\textbf{CHQ}} \\
BPE  & 2.94 & 40.59 & \textbf{45.63} & 33.77 & -  \\ 
\adapt & \textbf{2.67} & \textbf{41.92} & 44.57 & \textbf{37.60} & - \\
\hline
\end{tabular}}
\caption{Performance evaluation of \adapt{} on \medvoc{} model with BART-Large as base model. We observe gains of $1.87\%$ in Rouge-L (\textbf{R-L}) and $0.9\%$ in Concept Score (\textbf{CSr}). Improvements wherever observed are statistically significant (t-test: p-value$<0.001$). In High-OOV settings (R-L\textsubscript{H-O}) we observe gains of $10.40\%$ and $3.30\%$ in long-form generation (\textbf{R-L\textsubscript{L-RS}} considering only EBM and BioASQ). Notably, \adapt~results in the fragment score (FragSr) drop of $13.20\%$ across datasets.}
\label{tab:medvoc}
\end{table}

\noindent{\bf Human Evaluation.} We randomly select $40$ test data points sampled uniformly from four summarization datasets and follow the annotation procedure as described in ~\cite{fabbri-etal-2021-summeval,balde2024medvoc} to get annotations of summaries across the dimensions of \textit{relevance}, \textit{coherence} (on a Likert scale of $1$ to $5$), and \textit{faithfulness} (binary). Each annotator was given $30$ minutes to evaluate $10$ summaries and was compensated at a rate of $8$ UK pounds per hour (see Appendix~\ref{appendix:human-eval-app} for more details). Figure~\ref{fig:human-evaluation} shows the human evaluation results where \adapt~generates more faithful summaries ($97.5\%$ vs. $77.5\%$ of summaries are faithful), and more relevant summaries, where $82.5\%$ of data points get a positive score ($\ge 4$) in Likert scale, as compared to $65\%$ in case of BPE for \medvoc{}.

\begin{figure}[t]
    \centering
    \includegraphics[scale=0.4]{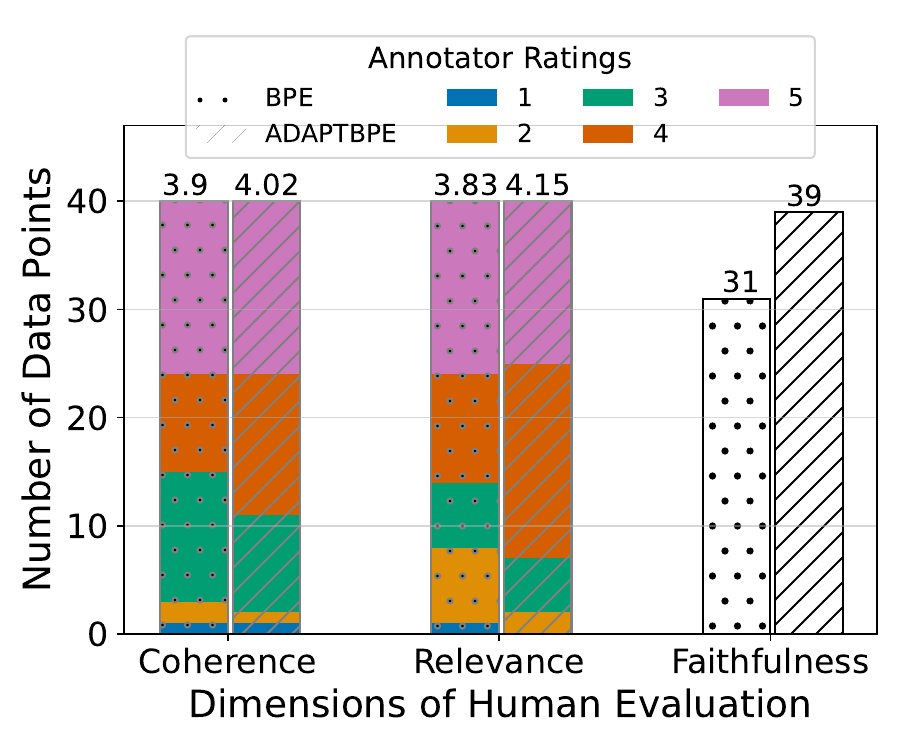}
    \caption{Human evaluation scores comparison over 40 randomly selected test data points. \adapt~produces more relevant, coherent, and faithful summaries during human evaluation with medical experts.}
    \label{fig:human-evaluation}
\end{figure}

\section{Conclusion}\label{sec:conclusion}
We are the first to show the incorrect BPE tokenization issue present in vocabulary adaptation techniques for fine-tuning PLMs to the target (expert) domain, designed for both classification and summarization tasks from various domains. The newly added target domain vocabulary is trivially added at low priority, causing BPE tokenizers to ignore them. Therefore, we propose a novel BPE tokenization scheme, \adapt{}, that modifies the BPE initialization step by searching through \vdomain{} to find the longest substring match. Our proposed \adapt{} algorithm is independent of the target domain-specific vocabulary construction algorithm and  focus only on improving the tokenization part. \adapt{}-enabled models outperform the competing baselines by $3.57\%$ and $1.87\%$ on average over classification and summarization tasks, respectively. Human evaluation using medical experts rate \adapt{}-based summaries to be more relevant and faithful than standard \bpe.

\section{Limitations}
We limit our evaluation to only pretrained language models and do not show results on large language models that also utilize BPE, such as LLaMa or Mistral, which uses Sentencepiece~\cite{kudo-richardson-2018-sentencepiece} Byte-level BPE Tokenization with fallback. We observe that  $27.76$\% of target domain-specific vocabulary terms are still tokenized into more than one subword (i.e., the ill-tokenization issue persists) for MEDVOC in the case of the LlaMa-2-7B model. However, the models considered in this study (BART and RoBERTa) use \textit{huggingface tokenizers} library~\cite{wolf-etal-2020-transformers} and we observed ill-tokenization in $64.13$\% of target domain-specific vocabulary terms. Thus, some efforts are needed to make \adapt{} work for LLMs. Second, the issue of ill-tokenization is mostly prevalent in the case of \bpe{} but less prevalent in the case of WordPiece tokenization, which is used by BERT and does not exist for Unigram tokenization scheme, which is used by PEGASUS and FLAN-T5 models. 

\section{Ethics Statement and Broader Impact}\label{sec:ethics}
Summarization and other text generation systems powered by large language models can suffer from hallucinations, producing outputs that deviate from the source material and are unfaithful summaries. While the proposed \adapt{} tokenization scheme generates more faithful summaries compared to existing baselines based on human evaluation, the summaries from such AI models are not yet reliable enough for high-stakes applications like medical contexts involving professionals and clinicians. Substantially more research is still needed to understand better the types of faithfulness and relevance errors made by these AI systems and to ultimately develop methods to mitigate or prevent such errors before these technologies can be safely deployed in sensitive real-world settings.

\bibliography{custom}

\appendix
\section{Experimental Setup}\label{appendix:expt-setup}

\subsection{Pre-trained Language Models}\label{appendix:model-type-desc}
To test the generalizability of our method described in Section~\ref{sec:methods}, we evaluate the efficacy of \drift{} on BART in case of summarization and RoBERTa in case of classification. %

\begin{itemize}
    \item \textbf{RoBERTa}~\cite{liu2019roberta}: RoBERTa (Robustly Optimized BERT Approach) is a transformer-based model, enhancing the original BERT model by training with more data and improved training techniques. It eliminates the Next Sentence Prediction (NSP) task used in BERT and employs dynamic masking during pre-training to increase robustness. RoBERTa is trained on a diverse corpus, including the Common Crawl dataset, to better capture nuanced language patterns. This model achieves state-of-the-art performance on various natural language processing (NLP) benchmarks. We use RoBERTa-base\footnote{\url{https://huggingface.co/FacebookAI/roberta-base}} which is a $125$ Million parameter model and uses Byte-pair Encoding tokenization case with a vocabulary ($|\text{V\textsubscript{PLM}}|$) of size $50265$.

    \item \textbf{BART}~\cite{lewis2020bart}: BART is a denoising autoencoder, implemented as a sequence-to-sequence model with a bidirectional encoder over corrupted text and a left-to-right auto-regressive decoder to generate the original document it was derived from. We use the BART-LARGE\footnote{\url{https://huggingface.co/facebook/bart-large}} model available from the \textit{huggingface} library. BART has $406$ Million parameters, uses \textit{Byte-Pair Encoding} tokenization, and its pretraining objective is a combination of \textit{Text Infilling} and \textit{Sentence Shuffling}. The vocabulary size of this PLM ($|\text{V\textsubscript{PLM}}|$) is $50265$. 
\end{itemize}

\subsection{Datasets}\label{appendix:datasets}
We describe here the details on the target task dataset mentioned briefly in Section~\ref{sec:expt-setup}.

\paragraph{Classification}
We use four target task datasets for classification that were used in \avocado{}. The dataset stats are described in Table~\ref{tab:dataset-stats-classification}. 
\begin{itemize}
    \item \textbf{CHEMPROT}~\cite{kringelum2016chemprot}. Chemprot dataset is a corpus used for the task of chemical-protein relation extraction. It consists of scientific abstracts annotated with various types of interactions between chemical compounds and proteins, such as inhibition, activation, and binding in total 13 classes. The dataset is commonly used to train and evaluate models in the domain of biomedical natural language processing, particularly for the extraction and classification of biochemical relationships.

    \item \textbf{ACL-ARC}~\cite{jurgens2018measuring}. The ACL-ARC dataset is designed to classify the intent behind citations in academic papers. It consists of annotated citations from research papers in the ACL Anthology, categorizing each citation based on its purpose, such as background, use, or comparison in total 6 classes. This dataset aids in understanding the functional and rhetorical roles of citations in scholarly communication.
    
    \item \textbf{HYPERPARTISAN}~\cite{kiesel2019semeval}. The hyperpartisan dataset consists of news articles labeled for hyperpartisanship, indicating whether they exhibit extreme bias. It was created to support research in detecting biased or partisan news content --a two-class classification, providing annotations on article-level and publisher-level partisanship. This dataset is used in natural language processing tasks to develop and benchmark models for identifying and understanding media bias.

    \item \textbf{AMAZON}~\cite{amazon-mcauley-2015}. The amazon dataset is a comprehensive collection of customer reviews and ratings from Amazon, covering a wide range of products. It includes detailed reviews, ratings, product metadata, and user information, providing a rich resource for sentiment analysis, recommendation systems, and other NLP tasks. The task is to identify whether a given review as input is actually helpful or not.
    
\end{itemize}
\begin{table}[!ht]
\centering
\scriptsize
\resizebox{\columnwidth}{!}{
\setlength{\tabcolsep}{0.1cm}
\begin{tabular}{ccccccc}
\hline
\textbf{Domain} & \textbf{Dataset} &  \multicolumn{3}{c}{\textbf{Document count}} &\textbf{Classes}  &\multicolumn{1}{c}{\textbf{OOV \%}} \\
 &  & \textbf{Train} &\textbf{Val} & \textbf{Test} &  & \textbf{RoBERTa} \\
\hline
BIOMED & CHEMPROT  & 4169 & 2427 & 3469 & relation (13) & 21.65 \\
CS & ACL-ARC & 1688 & 114 & 139  & citation intent (6) & 12.56\\
NEWS & HYPERPARTISAN & 515 & 65 & 65 & partisanship (2) & 3.94 \\
REVIEWS & AMAZON & 115251 & 5000 & 25000 &  helpfulness (2) & 3.69  \\
\hline
\end{tabular}}
\caption{Dataset statistics of downstream classification datasets. \textit{OOV\%} refers to the median fraction of unigrams in SD that are absent from the \base{} vocabulary.} %
\label{tab:dataset-stats-classification}
\end{table}

\paragraph{Medical Summarization}
We use four target task datasets in this study: two query-focussed summarization datasets, EBM and BioASQ, and two recent benchmark medical question summarization datasets, MeQSum and CHQSum, each of which we describe below.
\begin{itemize}
    \item \textbf{EBM}~\cite{DBLP:conf/acl-alta/MollaS11a}. Here input to the system is a query along with a PubMed abstract, and the expected output is the summary answering the question with the PubMed Abstract as the context. %

    \item  \textbf{BioASQ}~\cite{cite:BioASQ}. We use the dataset from BioASQ-9B Phase-B summarization task. The input to the system is a question followed by relevant snippets from a collection of PubMed Abstracts. There are two kinds of outputs an exact answer and an ideal answer associated with the input. For the summarization task, we consider the ideal answer as the Reference summary. %

    \item \textbf{MeQSum}~\cite{ben-abacha-demner-fushman-2019-summarization}. The dataset is created for better medical question summarization because the original patients’ questions are verbose. The dataset contains $1000$ patients’ health questions selected from a collection distributed by the U.S. National Library of Medicine. Each question is annotated with a summarized question by medical experts.

 \item \textbf{CHQSum}~\cite{yadav2022chq}. CHQSum consists of $1507$ domain-expert annotated question-summary pairs from the Yahoo community question answering forum\footnote{\url{https://webscope.sandbox.yahoo.com/catalog.php?datatype=l&did=11}} which provides community question answering threads containing users’ questions on multiple diverse topics and the answers submitted by other users. The authors with the help of $6$ domain experts identified valid medical question from the forum and asked the experts to formulate an abstractive summary for the questions. 
 
\end{itemize}
 

\begin{table}[!ht]
\centering
\scriptsize
\resizebox{\columnwidth}{!}{
\begin{tabular}{cccccccc}
\hline
\textbf{Dataset} &  \multicolumn{3}{c}{\textbf{Document count}} & \multicolumn{2}{c}{\textbf{Word count}} &\multicolumn{1}{c}{\textbf{OOV \%}} \\
{} & \textbf{Train} &\textbf{Val} & \textbf{Test} & \textbf{SD} & \textbf{RS} & \textbf{BART} \\ 
\hline
EBM & 1423 & 209 & 424  & 298  & 58  & 11.5 \\ 
BioASQ &  1525 & 491 & 496 & 505  & 40 & 9.4 \\ 
MeQSum &  700 & 150 & 150 & 70 & 12 & 5.7 \\ 
CHQSum & 1000 &  107& 400 & 184 & 12 & 6.3 \\ 
\hline
\end{tabular}}
\caption{Dataset statistics of downstream medical summarization datasets. \textit{OOV\%} refers to the median fraction of unigrams in RS that are absent from the \base{} vocabulary.} %
\label{tab:dataset-stats-summary}
\end{table}

\subsection{Evaluation Metrics} \label{appendix:eval-metrics}
We first describe the implementation details for computing Rouge scores discussed in Section~\ref{sec:expt-setup}, where we use the official Rouge~\cite{lin-2004-rouge} script\footnote{\url{https://github.com/bheinzerling/pyrouge/tree/master}}. The following parameters: \textit{-c 95 -2 -1 -U -r 1000 -n 4 -w 1.2 -a}, are used and we report the median at a $95$\% confidence interval. Additionaly, we also use Concept Score which identifies the medical conocept overlaps between generated and reference summary. To identify concepts, we use matcher.match utility of QuickUMLS~\cite{soldaini2016quickumls} tool in default setting.

\subsection{Added Vocabulary Sizes}~\label{appendix:added-vocab-sizes}
We mention the size of added vocabulary obtained by \avocado{} and \medvoc{} on classification and summarization datasets in Table~\ref{tab:added-vocab-stats}.

\begin{table}[h]
    \centering
    \scriptsize
    \begin{tabular}{lr}
     \hline
     \textbf{Dataset}        & $|\text{\vdomain{}}|$ \\
     \hline
     \multicolumn{2}{c}{{\bf \avocado{}} ($|\text{V\textsubscript{PLM}}|$: 50265)} \\
     CHEMPROT       & 5103 \\
     ACL-ARC        & 3419 \\
     AMAZON         & 1168 \\
     HYPERPARTISAN  & 743 \\ \hline
     \multicolumn{2}{c}{\textbf{\medvoc{}} ($|\text{V\textsubscript{PLM}}|$: 50265)} \\
     EBM            & 11061 \\
     BioASQ         & 6462 \\
     MeQSum         & 747 \\
     CHQSum         & 680 \\ \hline
     
    \end{tabular}
    \caption{Size of added vocabulary ($|\text{\vdomain{}}|$) for \avocado{}(RoBERTa) and \medvoc{}(BART) on classification and summarization datasets respectively.}
    \label{tab:added-vocab-stats}
\end{table}

\subsection{Hyperparameters}\label{appendix:hyperparameters}
We discuss the following hyperparameters: (i) the training hyperparameters, (iii) inference hyperparameters for MEDVOC.

\subsubsection{Training Hyperparameters}
\paragraph{\avocado{}.} All \avocado{} related experiments were run on one V100 32 GB graphic card. We kept the training hyperparameters same as that of what authors follow in the study. In brief, we tune learning rate : $\in \{1e-5, 2e-5, 5e-5\}$  and temperature: from $1.5$ to $3.5$ in steps of $0.5$.

\paragraph{MEDVOC.} All the experiments are run on one A100 40 GB GPU. We use the fine-tuning summarization scripts for BART provided in MEDVOC's codebase. We used the following hyperparameters to train BART model. learning rate: 5e-5, batch size: 32, and gradient accumulation steps: 8, rest all the hyperparameters takes its default values. We checkpoint at every 500 steps and train the model for a total of 5 epochs (approx 15K steps). The training times for IFT-PAC for \drift{} is mentioned in Table~\ref{tab:runtime-IFTS}.

\begin{table}[h]
    \centering
    \begin{tabular}{cc} \hline
       \textbf{Dataset} &\textbf{BART}  \\ \hline
       EBM      & 27 hrs 51 mins\\
       BioASQ   & 28 hrs 25 mins\\
       MeQSum   & 28 hrs 49 mins\\
       CHQSum   & 28 hrs 38 mins \\ \hline
    \end{tabular}
    \caption{Time required in hours for intermediate fine-tuning  using PAC for each target task dataset using BART with \adapt.}
    \label{tab:runtime-IFTS}
\end{table}

\subsubsection{Inference Hyperparameters}
We used beam search to run the \textbf{inference} on the test set. We tuned the following hyperparameters of beam search: beam size (B $\in [2,10]$) and length-penalty~\cite{wu2016googles} ($lp \in [0.1,3]$) on the validation split of the target task dataset. The best values of hyperparameters thus obtained are mentioned in Table~\ref{tab:generation_hyperparameters}.
 
\begin{table}[!ht]
    \centering
    \begin{tabular}{ccc} \hline
         \textbf{Dataset} &  $B$ & $lp$ \\ \hline
         EBM &  3 & 0.8\\
         BioASQ  &  6 & 0.8 \\
         MeQSum  &  8 & 0.1 \\
         CHQSum &  6 & 0.5 \\ \hline
    \end{tabular}
    \caption{Optimal values for inference hyperparameters - beam size ($B$) and Length Penalty ($lp$) used for beam-search generation for each of the datasets using BART with \adapt.}
    \label{tab:generation_hyperparameters}
\end{table}

\section{Human Evaluation}\label{appendix:human-eval-app}

Twelve individuals took part in an annotation task on the Prolific platform. Each person was asked to annotate ten random pairs of summaries from a pool of forty, with the order and source of the summaries concealed. Participants had thirty minutes to finish the task and were paid 8 UK Pounds per hour for their time. They also provided feedback on the experience and demographic information, excluding any personal details beyond what is made available by the platform. The task was conducted using Google Forms, with participants being shown a consent notice beforehand. 

\paragraph{Participation Criteria.} The filtering criteria for participants were kept same as that of MEDVOC~\cite{balde2024medvoc}: 
\begin{itemize}
    \item \textbf{Age:} $\ge 25$,
    \item \textbf{Primary Language:} English,
    \item \textbf{Highest education level completed:} Graduate degree (MA/MSc/MPhil/other), Doctorate degree (PhD/other)
    \item \textbf{Subject:} Medicine, Health and Medicine, Biomedical Sciences.
\end{itemize}

\paragraph{Annotation Guidelines.} The annotations were carried across three dimensions~\cite{fabbri-etal-2021-summeval} of coherence, relevance, and factual consistency. \textbf{Coherence} judges how well formed the summaries are and whether the sentences in the summaries are actually related to each other or not. \textbf{Relevance} judges how informative the summaries are considering the input as the context for evaluating relevance. \textbf{Factual Consistency} judges whether the facts, figures, numbers stated in the generated summary ca be verified from source input or not. Even if the generated text contains correct fact, but cannot be verified by only looking at input it is deemed as factually incosistent.

For each of these dimensions, we show one positive (high rating) and one negative example (low rating) along with an explanation  as a part of our annotation guideline (Table~\ref{tab:sample-annot-guideline}).

\begin{table}
    \scriptsize
    \centering
    \resizebox{\columnwidth}{!}{
    \begin{tabular}{p{1.27cm}p{6.5cm}}
    \hline
         \textbf{Source Document}& GE: Question in laymen terms: Has any genetic or other correlation ever been made between these two diagnosis? \\
                                & My 59 y.o. sister has a diagnosis of Periventricular Heterotopia.  Her 30 y.o. daughter has been suffering with same for last 15 years.  Her 37 Y.O. daughter is clinically full-care retarded (since infancy) and has severe idiopathic scoliosis.  I have severe idiopathic scoliosis. I use the term "severe" to express debilitating and multiple fusion surgeries.  All four of my generation female siblings have a level of scoliosis. FYI: this PH sister died last week, her remains are at the [LOCATION]\\ \hline
         \multicolumn{2}{c}{\textbf{Positive Example}}\\ \hline
         \textbf{Summary} & Can there be a genetic link between Periventricular Heterotopia and scoliosis?\\
         \textbf{Relevance} & $5$\\
         \textbf{Coherence} & $5$ \\ 
         \textbf{Factual Consistency} & $1$\\ \hline
         \textbf{Explanation} & Here we can see the summary is focused on idenitfying whether a genetic link exists b/w Periventricular Heterotopia and scoliosis which is what the user is asking about. \\ \hline
         \multicolumn{2}{c}{\textbf{Negative Example}}\\ \hline
         \textbf{Summary} & What are the causes of severe idiopathic scoliosis?\\
         \textbf{Relevance} & $1$\\
         \textbf{Coherence} & $1$ \\
         \textbf{Factual Consistency} & $0$\\ \hline
         \textbf{Explanation} & The question is asking for treatments of scoliosis which is not the theme of the input document.\\ \hline
         
    \end{tabular}}
    \caption{A negative and positive example as shown to the participant in the annotation guidelines for clarification under the three dimensions of annotation. The data point is taken from MeQSum dataset.}
    \label{tab:sample-annot-guideline}
\end{table}

\paragraph{Demographic analysis of participants.} The average age of participants was $29$ years. Out of $12$ participants, $10$ were female and  $2$\% were male. All the participants are Graduate studtents. The participants were recruited by platfrom from 3 countries: South Africa(3), Sweden(2), and UK(7).

\paragraph{Instruction on platform.} Prolific begins the user study with a clear instruction window describing what the task is about and what the participant is expected to do in the study. We attach the screenshot of that window which is shown to the participants in Figure~\ref{fig:instruction-snap}.

\begin{figure*}[!ht]
    \centering
    \includegraphics[width=\textwidth]{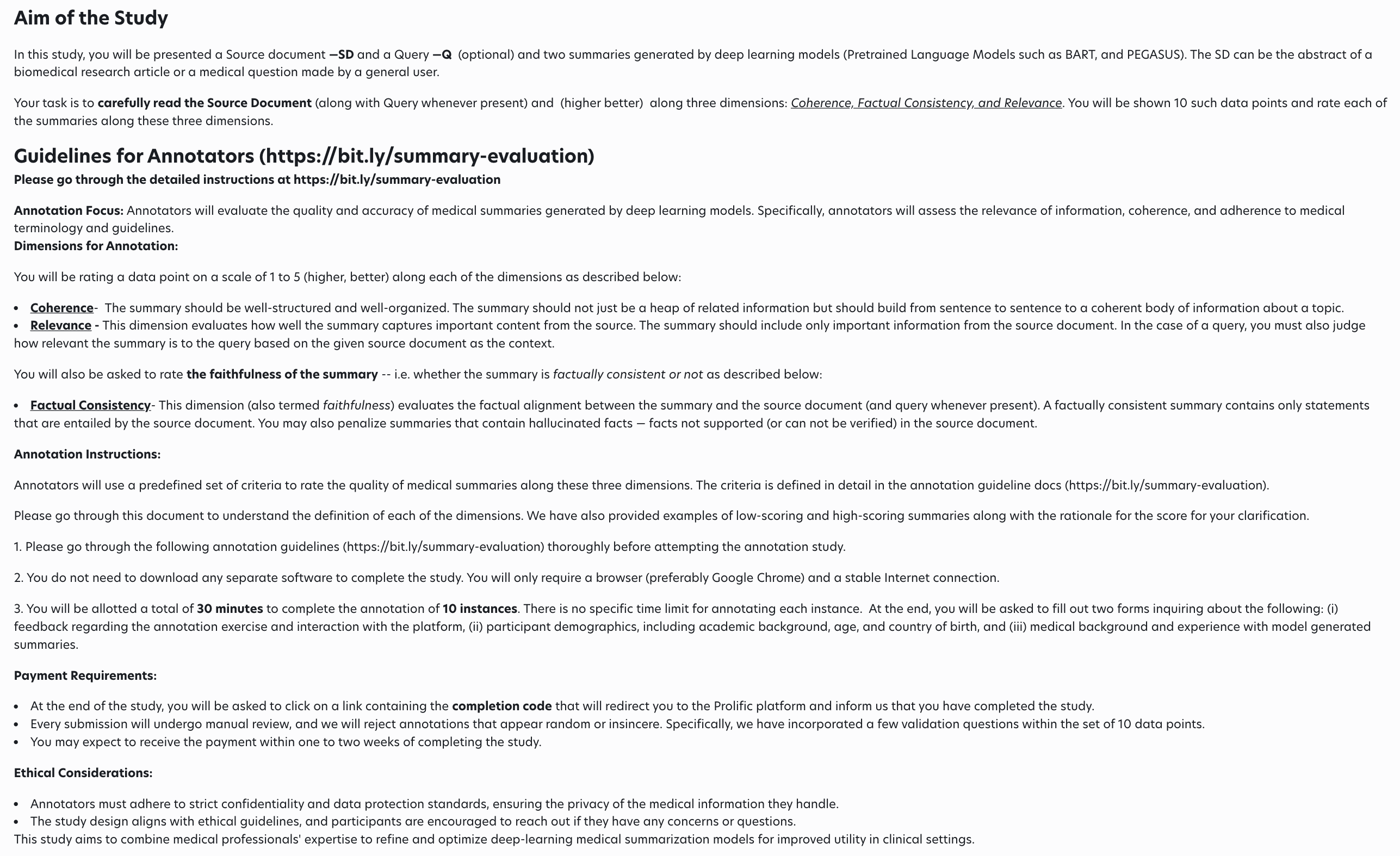}
    \caption{Instruction window as seen by an annotator participating in the study.}
    \label{fig:instruction-snap}
\end{figure*}
\end{document}